\documentclass[letterpaper]{article} 
\usepackage{aaai24}  
\usepackage{times}  
\usepackage{helvet}  
\usepackage{courier}  
\usepackage[hyphens]{url}  
\usepackage{graphicx} 
\usepackage{natbib}  
\usepackage{caption} 
\usepackage{algorithm}
\usepackage{algorithmic}
\newcommand{\mySymbolD}{\text{$\mathcal{D}$}}
\newcommand{\mySymbolC}{\text{$\mathcal{C}$}}
\newcommand{\mySymbolS}{\text{$\mathcal{S}$}}

\usepackage{makecell}
\usepackage{multirow}
\usepackage{color}
\usepackage{amsmath}
\usepackage{amssymb}
\usepackage{bbding}
\usepackage{lipsum}
\usepackage{tabularx}
 
\usepackage{array}
\usepackage[caption=false,font=footnotesize,labelfont=rm,textfont=rm]{subfig}
\usepackage{textcomp}

\usepackage{newfloat}
\usepackage{listings}
\DeclareCaptionStyle{ruled}{labelfont=normalfont,labelsep=colon,strut=off} 
\floatstyle{ruled}
\newfloat{listing}{tb}{lst}{}
\floatname{listing}{Listing}
\title{UFDA: Universal Federated Domain Adaptation with Practical Assumptions}
\author{
    Xinhui Liu\textsuperscript{\rm 1, \rm 2},
    Zhenghao Chen\textsuperscript{\rm 2},
    Luping Zhou\textsuperscript{\rm 2},
    Dong Xu\textsuperscript{\rm 3},
    Wei Xi\textsuperscript{\rm 1}\thanks{Wei Xi is the Corresponding Author.}, \\
    Gairui Bai\textsuperscript{\rm 1},
    Yihan Zhao\textsuperscript{\rm 1},
    Jizhong Zhao\textsuperscript{\rm 1} \\
}
\affiliations{
    \textsuperscript{\rm 1}School of Computer Science and Technology, Xi’an Jiaotong University, Xi’an, China\\
    \textsuperscript{\rm 2}School of Electrical and Computer Engineering, The University of Sydney, Sydney, Australia\\
    \textsuperscript{\rm 3}Department of Computer Science, The University of Hong Kong, Hong Kong SAR, China\\
    \texttt{liuxinhui@stu.xjtu.edu.cn, zhenghao.chen@sydney.edu.au, xiwei@xjtu.edu.cn}
}
 
\usepackage{bibentry}

\begin{document}
\maketitle
\begin{abstract}
Conventional Federated Domain Adaptation (FDA) approaches usually demand an abundance of assumptions, which makes them significantly less feasible for real-world situations and introduces security hazards. 
This paper relaxes the assumptions from previous FDAs and studies a more practical scenario named Universal Federated Domain Adaptation (UFDA). It only requires the black-box model and the label set information of each source domain, while the label sets of different source domains could be inconsistent, and the target-domain label set is totally blind. 
Towards a more effective solution for our newly proposed UFDA scenario, we propose a corresponding methodology called Hot-Learning with Contrastive Label Disambiguation (HCLD). It particularly tackles UFDA's domain shifts and category gaps problems by using one-hot outputs from the black-box models of various source domains.  
Moreover, to better distinguish the shared and unknown classes, we further present a cluster-level strategy named Mutual-Voting Decision (MVD) to extract robust consensus knowledge across peer classes from both source and target domains. Extensive experiments on three benchmark 
datasets demonstrate that our method achieves comparable performance for our UFDA scenario with much fewer assumptions, compared to previous methodologies with comprehensive additional assumptions.
\end{abstract}

\section{Introduction}
Federated Learning (FL)~\cite{mcmahan2017communication,mohassel2017secureml,mohassel2018aby3} allows models to be optimized across decentralized devices while keeping data localized, where no clients are required to share their local confidential data with other clients or the centralized server. 
Traditional FL often struggles to produce models that can effectively generalize to new unlabeled domains from clients due to the barrier presented by domain shifts~\cite{yang2019federated}. 
To address this, Federated Domain Adaptation (FDA)~\cite{fantauzzo2022feddrive,gilad2016cryptonets} are proposed and achieved tremendous success as it allows knowledge transfer from decentralized source domains to an unlabeled target domain using Domain Adaption (DA) techniques. 
Nonetheless, current FDA scenarios often operate under the presumption that model parameters or gradients are optimized based on the source domain. However, acquiring such information in real-world situations is exceptionally challenging due to commercial confidentiality.
Also, exposing such information introduces potential risks such as model misuse and white-box attacks. 

\begin{figure}[t]
\centering
\includegraphics[width=0.4\textwidth]{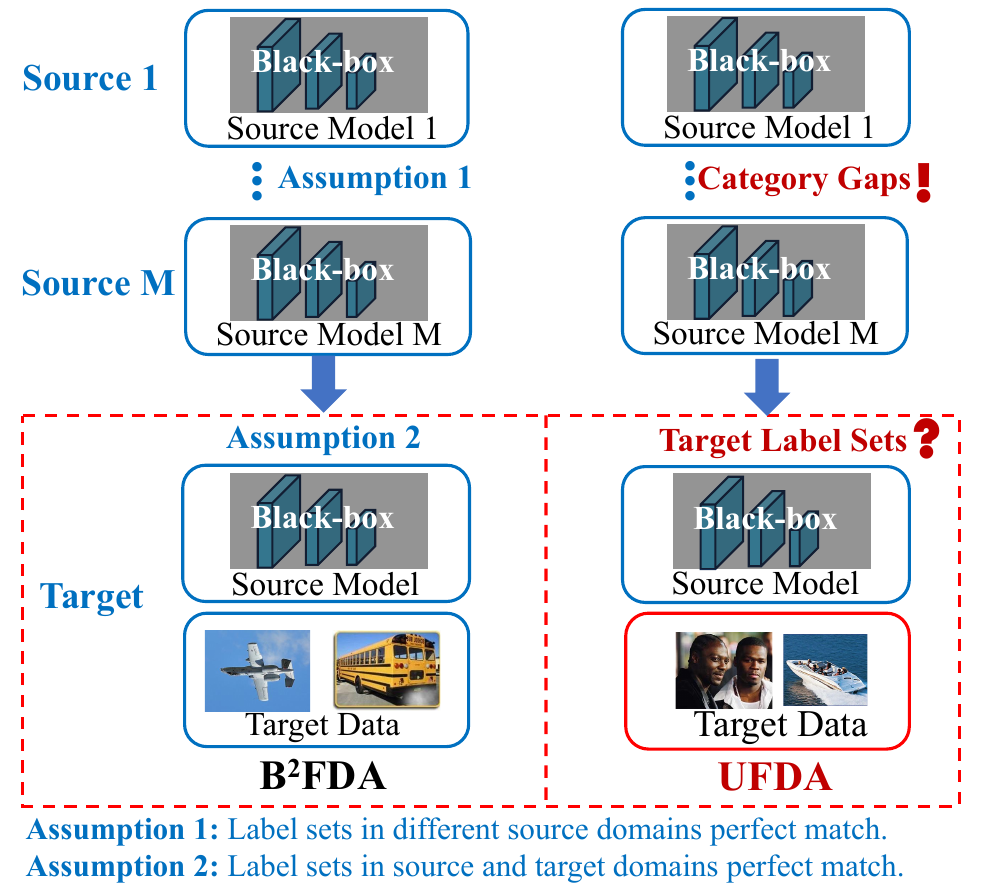}  
\caption{Overview of Federated Domain Adaptation for Black-Box Models (\textbf{B$^2$FDA}) (left), and our proposed Universal Federated Domain Adaptation (\textbf{UFDA}) (right). 
Different than B$^2$FDA, where the label set 
consistency among different source domains (\emph{i.e.}, \textbf{Assumption 1}) and between source and target domains (\emph{i.e.}, \textbf{Assumption 2}) are required, our UFDA scenario allows the label set diversity of source domains and the target domain.}
\label{task}
\end{figure}

To establish a relaxed condition, Federated Domain Adaptation with Black-Box Models (B$^2$FDA)~\cite{wu2021black,liang2022dine,10128163} is introduced,  where the target-domain client can only access the application programming interfaces (APIs) of various source domains. However, most existing B$^2$FDA approaches assume that the label sets of different source domains must perfectly align with each other and that of the target domains. This assumption is particularly challenging to fulfill in real-world scenarios.
First, source data can originate from vastly diverse domains. For example, the biometric data of a single client could stem from unrelated sources like the medical domain (\emph{e.g.}, clinical records from different hospitals) or the financial domain (\emph{e.g.}, user records from different Banks). Second, in real-world scenarios, acquiring information about the label set of the target domain samples is often a formidable task. Consequently, attempting to align the label sets of source and target domains becomes impractical.

To further minimize those in-practical assumptions from B$^2$FDA, we introduce a new scenario Universal Federated Domain Adaptation (UFDA) towards a practical FDA set-up with practical assumptions. 
As shown in Figure.~\ref{task}, in UFDA, the target domain solely requires a black-box model, devoid of its specifics (\emph{e.g.}, gradients). Meanwhile, we only need to know the source domains' label sets, which are not required to be identical as in B$^2$FDA scenarios, and the target domain's label set will remain entirely unknown as most real-world DA scenarios.
On the other hand, different from most existing B$^2$FDA setups~\cite{liang2022dine}, our UFDA presents two unique challenges:
\underline{First}, as the target domain's label set is completely unknown, the model optimized based on each individual source domain could be particularly imprecise for those unique categories of the target domain. 
\underline{Second}, the completion uncertainty of the target domain's label set also makes it impossible to distinguish the shared and unknown classes among source and target domains. However, it is important in FDA problems to guarantee the consistency of label sets between source and target domains. 

To tackle the first challenge, we propose a methodology called Hot-Learning with Contrastive Label Disambiguation (HCLD). 
It adopts one-hot outputs (without confidence) produced by various source APIs, which generate more than one candidate pseudo-labels for each target sample. Compared with previous FDA methods, which directly adopt one candidate (with confidence) from source APIs by using the probability function (\emph{e.g.}, Softmax), our method can mitigate the impact caused by the falsely higher confidence in these non-existent categories. 
To obtain more credible pseudo-labels, we propose a Gaussian Mixture Model (GMM) based Contrastive Label Disambiguation (GCLD) method, which sharpens the shared-class confidence and smooths the unknown-class confidence. Specifically, it leverages contrastive learning (CL)~\cite{khosla2020supervised} strategy to dynamically generate prototype-based clustering, which will fit a GMM~\cite{permuter2006study} based on its self-entropy distribution for sample divisions. 
Therefore, the easy-to-learn sample can be treated as a shared-class sample while the hard-to-learn sample can be treated as an unknown-class sample. 
Furthermore, to address the second challenge, we propose a cluster-level Mutual-Voting Decision (MVD) strategy by leveraging the consensus knowledge of shared classes among source and target domains. 
We calculate a ``mutual voting score" for each class based on the overlapping samples recognized as the same category from all APIs (\emph{i.e.}, source + target). Then, we use this score to distinguish each class as ``shared" or ``unknown" type.

Our contributions are summarized as follows:
\begin{itemize}
\item We introduce a new FDA scenario, UFDA, which not only inherits relaxed assumptions as in B$^2$FDA, but also eliminates the consistency requirement of label sets among source domains and keeps the target domain's label sets completely unknown, towards a practical scenario for real-world situations.
\item We proposed a novel methodology, HCLD, to address the imprecision issue for samples from non-existent categories. It adopts ensemble one-hot outputs from multi-source APIs to produce multiple candidate pseudo-labels and uses a GMM-based strategy GCLD to disambiguate those candidates.
\item We present a cluster-level strategy MVD to distinguish shared and unknown classes by leveraging consensus knowledge across peer classes from source and target domains. 
\item We conduct extensive experiments on three DA benchmarks. The results demonstrate that our method exhibits performance on par with previous MDA approaches, yet relies on significantly fewer assumptions. This substantiates the practicality of our method.
\end{itemize}
  
\begin{figure*}[!ht]
\centering
\includegraphics[width=0.95\textwidth]{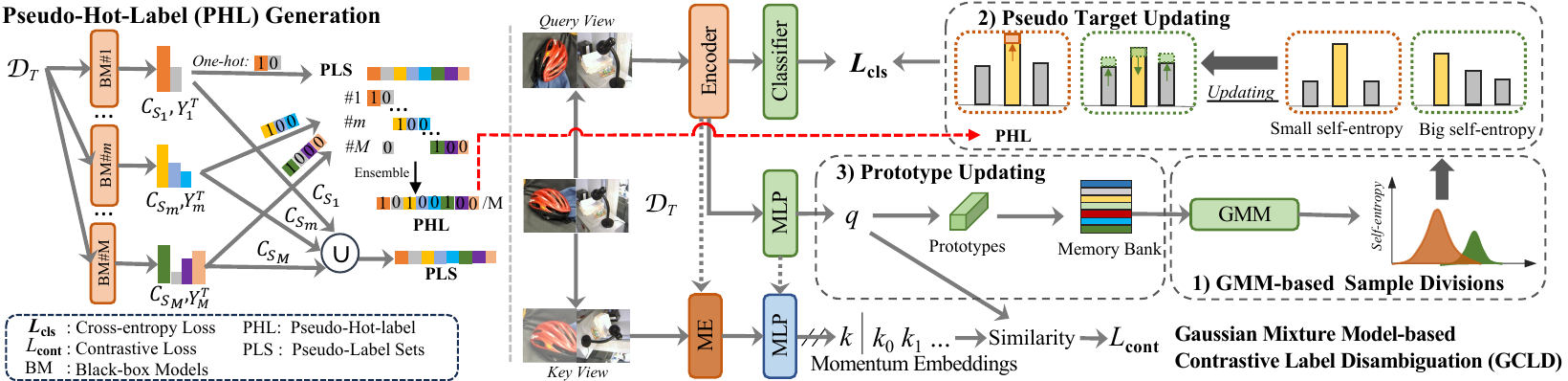} 
\caption{Our Hot-Learning with Contrastive Label Disambiguation (HCLD) methodology contains two key components: Pseudo-Hot-Label (PHL) Generation and Gaussian Mixture Model-based Contrastive Label Disambiguation (GCLD). $\mySymbolD_T$, ME, `$//$’, GMM, and MLP respectively represent the unlabeled target dataset, Momentum Embeddings maintained by a queue structure, the stop-gradient operation, Gaussian Mixture Model, and the Multi-Layered Perceptron module. }
\label{model} 
\end{figure*} 

\section{Related Works}
\subsection{Multi-source Domain Adaptation (MDA)}
MDA~\cite{ben2010theory,blitzer2007learning,liu2022m2n} has gained significant attention as a means to mitigate performance degradation caused by domain shifts. 
Despite the achievements of MDA, many existing approaches~\cite{hoffman2018algorithms,zhao2018adversarial,chen2017neural,chen2022exploiting} are limited to the assumption of perfectly matched label sets and have to access the raw multi-source, which can be inefficient and may raise concerns regarding data protection policies~\cite{voigt2017eu}. 

To tackle category shift issues, the UniMDA scenario is introduced~\cite{xu2018deep,kundu2020universal,shui2021aggregating,shui2022novel,saito2021ovanet,shan2023prediction,chen2022evidential}, where the label set among multi-sources differ, and no prior knowledge about the target label sets is accessible. 
In UniMDA, the concept of category shift was first introduced in DCTN~\cite{xu2018deep}, which acknowledged that the number of categories in each source domain may differ from the target domain. 
DCTN learns transferable and discriminative representations via an alternating adaptation algorithm and a distribution-weighted combining rule. 
To address data privacy issues, source-free domain adaptation (SFDA)~\cite{liang2020we,ahmed2021unsupervised,dong2021confident,9530705,zhao2022source} and federated domain adaptation (FDA)~\cite{li2020multi,peng2019federated,feng2021kd3a,wu2021collaborative} have attracted increasing attention.
Instead of accessing the raw data directly, SFDA utilizing the well-trained model rather than the raw labeled data has emerged as a possible solution to this problem. 
Another setting that deals with unavailable source data is the FDA, where the goal is to develop a global model from decentralized datasets by aggregating the parameters of each local client~\cite{csurka2022visual}. Inspired by FL, ~\cite{peng2019federated} first raised the concept of the FDA. This work provides a solution named Federated Adversarial Domain Adaptation, which aims to address the FDA problem in a federated learning system using adversarial techniques. 

However, these approaches do not address both of these limitations simultaneously. Recently, a few works ~\cite{kundu2020universal,saito2020universal,qu2023upcycling} explore SFDA under category shift. 
Despite their effectiveness, they require dedicated multi-source model specifics, which can be restricted due to their commercial value and associated risks, such as model misuse and white-box attacks. 
In this work, we deal with a practical scenario of UFDA, which requires neither the shared
data and model specifics, consistency of label sets among source domains, and information on the target domain label set. 

\subsection{Contrastive learning (CL)} 
Due to the success of CL~\cite{wu2018unsupervised,chen2020simple,chen2021improving,chen2022lsvc,guo2023bridging}, numerous efforts have been made to improve the robustness of classification tasks by harnessing the advantages of CL. For instance, ~\cite{zheltonozhskii2022contrast} employed CL as a pre-training technique for their classification model. Another approach, RRL~\cite{li2021learning} introduced label cleaning utilizing two thresholds on soft labels, which are calculated from the predictions of previous epochs and their nearest neighbors. Similarly, Sel-CL~\cite{li2022selective} leveraged nearest neighbors to select confident pairs for supervised CL~\cite{khosla2020supervised}. Despite their demonstrated effectiveness, these methods are not explicitly designed to tackle the category shift between the noise-label sets and the ground-truth label set. 
 
\section{Methodology}  
\subsection{Preliminaries} 
We are given $M$ source datasets from different clients $\{\mySymbolD_S^{m}\}_{m=1}^{M}$ and an unlabeled target client $\mySymbolD_T$, where each source client contains $N_m$ labeled source samples $\mySymbolD_{S}^{m}:=\{(x_{i}^{m},y_{i}^{m})\}_{i=1}^{N_{m}}$ and the target client comprising $N_T$ unlabeled samples $\{x_{i}\}_{i=1}^{N_{T}}$, $s.t., {x_{i}}\in{X^{T}}$.  
In most real-world scenarios, each client's data and model specifics are stored exclusively on local systems, ensuring that they are not shared with other clients or a centralized server. Therefore, the label sets between the aforementioned multi-source and target clients may exhibit significant variations. 

While, most existing FDA studies intuitively assume that multi-source and target clients share the same label sets, which is not practical. Inspired by the research of UniMDA, we define $\mySymbolC_{s_{m}}$ as the label sets for the $m$-th source node and $\mySymbolC_{t}$ as the label set for the target node. The label sets $\mySymbolC_{m}$ represents the common labels between $\mySymbolC_{s_{m}}$ and $\mySymbolC_{t}$. Furthermore,
$\overline{\mySymbolC}_{s_{m}}=\mySymbolC_{s_{m}} \backslash \mySymbolC_{m}$ represent the label sets exclusive to $\mySymbolD_{S}^{m}$. Similarly, $\overline{\mySymbolC}_{t}=\mySymbolC_{t} \backslash \{{\cup}_{m}\mySymbolC_{m}\}$ indicates the classes in the target domain $\mySymbolD_T$ that are unknown in the multi-source domains, as they should never appear in any source label sets. 
The label sets $\mySymbolC$ represent the union of shared classes, \emph{i.e.}, $\mySymbolC = {\cup}_{m}\mySymbolC_{m}$. It is important to note that the target data are fully unlabeled and the target label set (which is inaccessible during training) is only used to define the UFDA problem. 

\subsection{HCLD}
Our proposed HCLD aims to establish an effective mapping that can accurately classify target samples if they correspond to the shared class $\mySymbolC$, or confuse the samples with an "unknown" class.   
As shown in Figure.~\ref{model}, HCLD consists of two key components: 1) Pseudo-Hot-Label (PHL) Generation;
2) Gaussian Mixture Model-based Contrastive Label Disambiguation (GCLD).
Firstly, to mitigate the impact caused by multi-source APIs' falsely higher confidence for the non-existent categories, we calculate the pseudo-labels for each target sample with the proposed PHL Generation strategy.  
Then, we adopt the GCLD manner to obtain more credible pseudo-labels, which sharpens the shared-class confidence and smooths the unknown-class confidence.  
  
\subsubsection{PHL Generation} 
In UFDA, only the label sets $\{\mySymbolC_{s_{m}}\}_{m=1}^{M}$ in each source domain and the softmax output $Y^{T}_{m}$ in each source APIs for target samples are acceptable for the target party: $Y^{T}_{m}  = f^{m}_{S}(X^{T})$. 
%
Considering the domain shift between multiple source and target domains, the key challenge lies in obtaining more reliable pseudo-labels for each target sample. Empirically, individual source APIs often display increased confidence levels for both shared- and non-existent categories. Such a trend adversely affects the accuracy of pseudo-labels that are produced using these confidence scores.

To address the aforementioned limitation, 
we suggest the use of an ensemble of multiple one-hot outputs to create the pseudo-labels, referred to as PHL $C_{pse}$ (\textit{i.e.,} Figure.~\ref{model}), which generates multiple candidate pseudo-labels for each target sample, providing a broader and potentially more accurate range of labeling options.
Given the lack of pre-existing knowledge about the target label sets, we determine the Pseudo-Label Sets (PLS) for the target domain by the following method:  $\hat{\mySymbolC}_{T}= {\cup}_{m}\mySymbolC_{s_{m}}$.  This strategy ensures that the accurate labels identified by each APIs are encompassed within these candidate pseudo-labels. 
  
\subsubsection{GCLD}
The candidate pseudo-labels in the above PHL $C_{pse}$ inevitably contain unknown categories due to the gap between multi-source and target domains. We adopt GCLD, which iteratively sharpens the possible shared-class confidence, smooths the possible unknown-class confidence, and obtains more credible pseudo-labels. 
 
The critical challenge is distinguishing between shared- and unknown-class samples.
Inspired by~\cite{permuter2006study}, GMM can better distinguish clean and noisy samples due to its flexibility in the sharpness of distribution. 
Treating easy-to-learn samples as shared class instances and challenging samples as unknown-class instances, we facilitate the acquisition of discriminative image representations through CL and construct a GMM over the representations for sample divisions. 
Typically, the dimension of contrastive prototypes is limited by the pseudo-label sets, making it difficult for GMM to handle this scenario effectively.
Therefore, we utilize a comprehensive Memory Bank denoted $U^e=\left\{u_1^e, \ldots, u_{N_T}^e\right\}$ that maintains the running average of the features of all target samples. Here, $U^e$ represents the Memory Bank in epoch $e$. We initialize $U^e$ with random unit vectors and update its values by mixing $U^e$ and ${U}^{e-1}$ during training (details in the next subsection).
\begin{equation}
U^e \leftarrow \delta {U}^e+(1-\delta) U^{e-1}
\end{equation}
where $\delta$ is a mixing parameter. The self-entropy of $U^e$ for each sample can be defined as:
\begin{equation}\label{memory_bank} 
l_{ce}(i) = -\sum u^e_i \log \left(u^e_i\right), i \in\{1, \ldots, N_T\}
\end{equation}

\textbf{1) GMM-based Sample Divisions.} To distinguish between shared- and unknown-class samples, we fit a two-component GMM to the self-entropy distribution $l_{ce}$ using the Expectation Maximization algorithm. 
Each sample is assigned a shared probability $w_{i}$, which is the posterior probability $p\left(\theta \mid \ell_{ce}\right)$, where $\theta$ corresponds to the Gaussian component with a smaller mean (indicating a smaller self-entropy). Based on the shared probability, we divide all target samples into two sets: $\textit{W}^{1}$ (the sample may with shared class) and $\textit{W}^{0}=D_{T} \backslash \textit{W}^{1}$ (the sample may with unknown class) by setting a threshold $\sigma$. 

\textbf{2) Pseudo Target Updating.}
In terms of the above distinguished shared class $\textit{W}^{1}$ and unknown class samples $\textit{W}^{0}$, we sharpen the shared-class confidence and smooth the unknown-class confidence to update the pseudo-labels $C_{pse}$ as follows,
\begin{equation}\label{updating_target0} 
{C}_{pse}^{e} \leftarrow \phi(\phi{C_{pse}^{e}} + (1-\phi){C}_{pse}^{e-1})+ (1-\phi)\boldsymbol{z}^{e}
\end{equation}
\begin{equation}\label{updating_target1} 
\quad \boldsymbol{z}^{e} = \begin{cases} Onehot(u^e_{i}) & x_i \in \textit{W}^{1} \\ 1/n_{C} & $otherwise.$
\end{cases}
\end{equation}
where $\phi$ is a tunable hyperparameter. Ultimately, the pseudo-labels with shared classes will be clustered around each cluster center, while confusing the pseudo-labels with unknown labels.

\textbf{3) Prototype Updating.}
Since the contrastive loss induces a clustering effect in the embedding space, we maintain a prototype embedding vector $\mu_c$ corresponding to each class in $\hat{\mathcal{C}}_{T}$, which serves as a set of representative embedding vectors. This approach adopts updating $\mu_c$ with a moving-average style:
\begin{equation}
\begin{aligned}
\boldsymbol{\mu}_c = &\operatorname{Normalize}\left(\gamma \boldsymbol{\mu}_c+(1-\gamma) \boldsymbol{q}\right),\\
& \quad \text { if } c=\arg \max _{j \in \hat{\mathcal{C}}_{T}} f^j\left(\operatorname{Aug}_q(\boldsymbol{x})\right)
\end{aligned}
\end{equation}
where the momentum parameter $\gamma$ was set as 0.99. Then, we iteratively update the above-mentioned Memory Bank $U^e$ with the moving-updating mechanism $U^e \leftarrow \boldsymbol{q} * \boldsymbol{\mu}_c^{T}$ (details of $\boldsymbol{q}$ in the next subsection).

\begin{table*}[!htbp]
\centering
\setlength{\tabcolsep}{5.9pt}
\begin{tabular}{l|c|c|c|c|c||c|c|c|c||c|c|c|c}
    \hline
    & \multicolumn{5}{c||}{\textbf{Office-Home}}& \multicolumn{4}{c||}{\textbf{Office-31}}& \multicolumn{4}{c}{\textbf{VisDA+ImageCLEF-DA}}\\
    \cline{2-14}\multirow{1} *{\textbf{Methods}} & \multirow{1}*{Ar} & \multirow{1}*{Cl}& \multirow{1} *{Pr} & \multirow{1} *{Re}  & \multirow{1}*{Avg.}&A & D & W  & Avg.& C & I& P & Avg.\\
    \hline
    DANN & 68.97 & 53.37 & 79.70 & 82.09 & 71.03 & 83.43 & 81.36& 84.22& 83.00  & 76.25 & 64.75 & 59.75 & 66.92 \\
    RTN & 68.72 & 59.97 & 77.04 & 86.00 & 72.93& 86.70 & 88.64 & 83.22& 86.19   & 81.50 & 67.75 & 62.75 &  70.67 \\
    OSBP & 44.17 & 45.98 & 63.37 & 68.56 & 55.52& 57.76 & 81.54& 78.48& 72.59   & 49.75 & 44.50 & 44.25 & 46.17\\
    UAN & 69.27 & 60.32 & 79.78 & 82.82 & 73.05 & 85.35 & 94.54& 92.03& 90.65   & 75.00 & 67.75 & 61.00 &  67.92\\
    DCTN & 64.77 & 42.09 & 65.25 & 70.11 & 60.56 & \underline{88.84}  & 89.18& 83.73 & 87.25   & 66.25 & 55.75& 50.50 & 57.50 \\
    MDAN & 67.56 & 55.36 & 79.20 & 86.02 & 72.04 & 85.82 & 92.82& 88.43 & 89.02  & 68.50 & 65.25& 60.50 & 64.75 \\
    MDDA & 44.66 & 34.54 & 54.93 & 53.24 & 46.84  & 84.91 & 89.16& 89.60 & 87.89 & 60.25 & 44.50&  36.50& 47.08  \\
    UMAN & \underline{79.00} & \textbf{64.68} & 81.12 & 87.08 & \underline{77.97} & \textbf{90.22} & 94.50& 94.53 & \textbf{93.08} & \textbf{88.00} & \textbf{83.25} & \textbf{70.50} &  \textbf{80.58} \\ 
    \hline 
    \multicolumn{14}{c}{Data, Model Specifics: Available$\uparrow$, Data, Model Specifics: Decentralized $\downarrow$} \\
    \hline
    \textbf{HCLD$^2\star$} &\textbf{80.31} & \underline{62.27} &\textbf{82.33} &\textbf{88.85} & \textbf{78.42} & 80.00 & \underline{96.55}& \textbf{96.52}& \underline{91.02} & 77.25  & 67.54& 60.05 & 68.28  \\
    \textbf{HCLD$^2$} &77.38 &61.09 & \underline{81.74} & \underline{88.49} & 77.18 &75.66 & \textbf{97.22} &\underline{94.95} & 89.28 & \underline{84.25} &\underline{73.50} & \underline{68.50} & \underline{75.41}  \\ 
    \hline
\end{tabular} 
\caption{Comparison with the State-Of-The-Art DA methods on three DA benchmarks (Backbone: Resnet-50) measured by Accuracy (\%). The best numbers are highlighted in \textbf{bold}. The second numbers are highlighted with \underline{underline}. Different from HCLD$^2$, HCLD$^2\star$ implements HCLD$^2$ with the Pseudo-Soft-Label.}
\label{officehome_ofice31} 
\end{table*}

\subsubsection{Training Objective}
Given the target samples with PHL $\{x^{i}, c_{pse}^{i}\}_{i=1}^{N_{T}}$, we generate a query view $Aug_{q}(x)$ and a key view $Aug_{k}(x)$ with the randomized data augmentation $Aug(x)$. Then, HCLD employs the query network $g(.)$ and the key network $g^{\prime}(.)$ to encode the query $\boldsymbol{q} = g(Aug_{q}(x))$ and keys $\boldsymbol{k} = g^{\prime}(Aug_{k}(x))$. Similar to MoCo~\cite{he2020momentum}, the key network employs a momentum update using the query network. Additionally, we maintain a queue that stores the most recent key embeddings $\boldsymbol{k} $ and chronologically update the queue. This enables us to establish a contrastive embedding pool $A = B_{q} \cup B_{k} \cup queue$, 
where $B_{q}$ and $B_{k}$ represent vectorial embeddings corresponding to the query and key views, respectively. For each sample, the contrastive loss can be calculated by contrasting its query embedding with the remaining embeddings in pool $A$. 
\begin{equation}\label{cont} 
\begin{aligned}
\mathcal{L}_{\text {cont }}&(g ; \boldsymbol{x}, \tau, A) = -\frac{1}{|P(\boldsymbol{x})|} \\
&\sum_{\boldsymbol{k}_{+} \in P(\boldsymbol{x})} \log \frac{\exp \left(\boldsymbol{q}^{\top} \boldsymbol{k}_{+} / \tau\right)}{\sum_{\boldsymbol{k}^{\prime} \in A(\boldsymbol{x})} \exp \left(\boldsymbol{q}^{\top} \boldsymbol{k}^{\prime} / \tau\right)},
\end{aligned} 
\end{equation} 
where $A(x) = A \backslash \{\boldsymbol{q}\}$ and $\tau \geq 0$ is the temperature parameter. Inspired by~\cite{shen2019learning}, DNNs first memorize the training data of easy-learning samples, then gradually adapt to noisy labels. We construct the positive set $P(x)$ with the predicted label from the Classifier (See Figure.~\ref{model}). About the query view, we train the classifier $f$ using cross-entropy loss, 
\begin{equation}\label{cls} 
\mathcal{L}_{\mathrm{cls}}\left(f; \boldsymbol{x}_{i}, c_{pse}^{i}\right) = \sum_{n=1}^{n_{C}}-s_{n}^{i} \log \left(f^n\left(\boldsymbol{x}_{i}\right)\right), x_i \in X^{T}
\end{equation}
where $n_{C}$ indicates the number of categories in $\hat{C}_{T}$, $n$ denotes the indices of labels, $s_{n}^{i}$ denotes the $n$-th vector of $c_{pse}^{i}$, and $f^{n}$ denotes the $n$-th output of $f$. Putting it all together, the overall loss function can be defined as $\mathcal{L}=\mathcal{L}_{\text {cls}}+\beta \mathcal{L}_{\text {cont}}$, where $\beta$ is set as 0.01 to balance each loss component.
 
\begin{figure}[!t]
\centering
\includegraphics[width=0.4\textwidth]{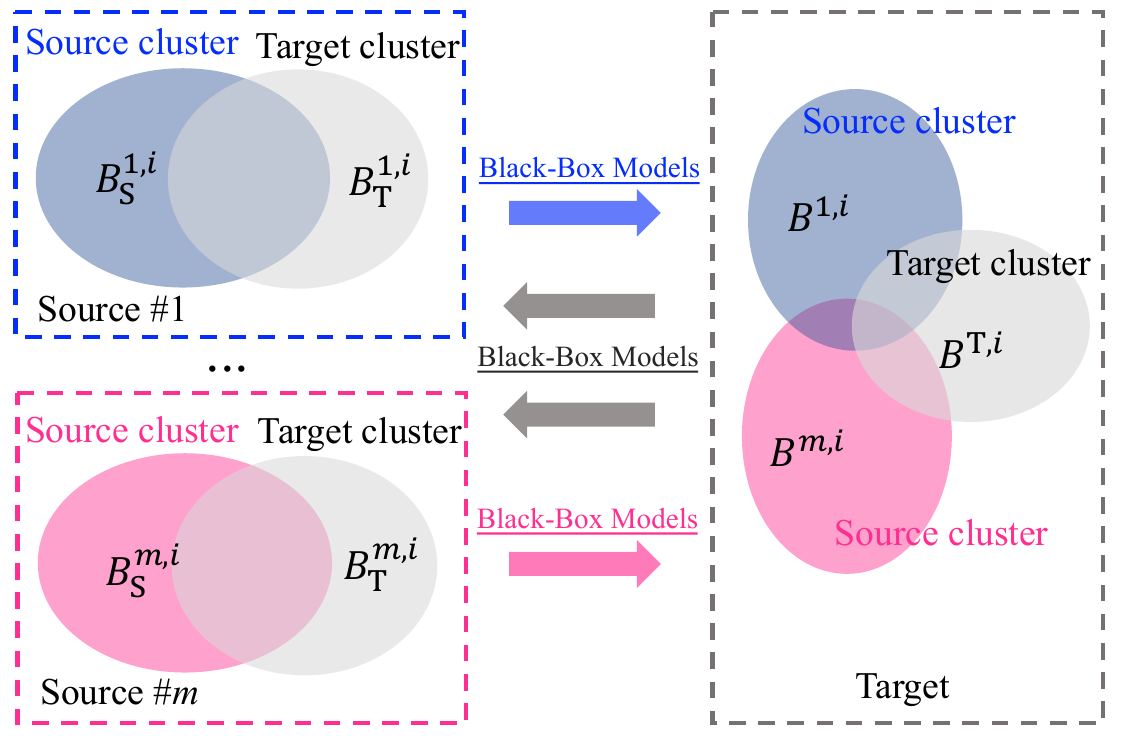} 
\caption{Overview of Mutual-Voting Decision (MVD).}
\label{mutual_voting}
\end{figure}

\subsection{MVD}
Through the above HCLD strategy, we could optimize a model that is well-performing in shared classes and ambiguous in unknown classes. 
However, better adaptation performance depends on accurate inference for shared and unknown classes, which becomes challenging without multi-source data or parameters. 
Inspired by the consensus knowledge of shared classes among different domains, we consider utilizing cluster-level consensus from multi-source and target APIs to distinguish between shared and unknown classes. 
As the source and target APIs rarely misunderstand the non-existent category as the same shared class, we introduce an MVD strategy, which leverages the knowledge voting among the source and target views. 
Specifically, it calculates the voting scores in each class (the proportion of overlapping samples recognized as the same category in the dataset by all APIs, compared to the minimum number of all samples recognized as that category among these APIs) and calculates the mutual voting scores among the source and target views, which can be used to determine if it reaches a consensus.
The overview of MVD is shown in Figure.~\ref{mutual_voting}.   
For the source view, given a pair of matched class clusters $B_S^{m, i}$ (obtain $B_S^{m, i}$ across the outputs of the $m$-th source model) and $B_T^{m, i}$, we measure the cluster-level consensus via calculating the voting score $\left\{d_s^1, \ldots,d_s^{n_{C}}\right\}$, 
\begin{equation}\label{source_0}
d_{s}^{i,m} = \frac{(B_S^{m,i}\cap B_T^{m,i})}{\arg \min _{j \in \{1, \ldots, M\}} (B_S^{j,i}, B_T^{j,i})}, i \in \mathcal{C}_{S_{m}}
\end{equation}
\begin{equation}\label{source_1} 
d_s^i = \arg \max _{j \in \{1, \ldots, M\}} d_{s}^{i,j}
\end{equation}
Similarly, we calculate the voting score $\left\{d_t^1, \ldots,d_t^{n_{\mySymbolC}}\right\}$ in the target view. Then, the mutual-voting score of two views for each union source class can be calculated as:
\begin{equation}\label{voting}
\mathcal{S}_{c} =\frac{d_t^c+ d_s^c}{2}, c \in \hat{\mathcal{C}}_{T}
\end{equation}
For each $\mySymbolS_{c}$, we can predict the class $c$ with a validated threshold $\lambda$. 
This either assigns class $c$ to one of the union source classes ($\mySymbolS_{c}\textless \lambda$) or rejects it as an "unknown" class. 

\section{Experiments}
\subsection{Experimental Setup}  
\subsubsection{Datasets.}
\textbf{Office-Home}~\cite{venkateswara2017deep} is a DA benchmark that consists of four domains: Art (Ar), Clipart (Cl), Product (Pr), and Real World (Re). \textbf{Office-31}~\cite{saenko2010adapting} is another popular benchmark that consists of three domains: Amazon (A), Webcam (W), and Dslr (D). \textbf{VisDA2017+ImageCLEF-DA} is a combination of two datasets. VisDA2017~\cite{peng2018visda} is a DA dataset where the source domain contains simulated images (S) and the target domain contains real-world images (R). ImageCLEF-DA, on the other hand, is organized by selecting the common categories shared by three large-scale datasets: ImageCLEF (C), ImageNet (I), and Pascal VOC (P). Classes in the combined dataset are numbered as follows: Classes No. 1–7 represent the shared classes among the five datasets in alphabetical order. Classes No. 8–12 are the remaining classes from S and R domains. Classes No. 13–17 are the remaining classes from the C, I, and P domains. 
In UFDA, each domain contains two types of labels: shared and unknown. We use a matrix to describe the specific UniMDA setting, called UMDA-Matrix~\cite{yin2022universal}, which is defined as $\left[\begin{matrix}
  |\mySymbolC_1| &...& |\mySymbolC_M|& |\mySymbolC| \\
   |\overline{\mySymbolC}_{S_1}|&...&|\overline{\mySymbolC}_{S_M}| & |\overline{\mySymbolC}_t| \\
  \end{matrix} \right]$.
The first row is the size of the shared class of all the domains, and the second row denotes the unknown class. The first $m$ columns are the label set of the multi-source domains, and the last one denotes the target domain. In this way, UniMDA settings can be determined by the division rule. To ensure a fair comparison with previous UniMDA works, we maintain the same UMDA-Matrix settings with UMAN. 

\subsubsection{Baseline Methods.}
The proposed HCLD$^2$ (HCLD \& MVD) is compared with a range of State-Of-The-Art (SOTA) DA approaches. \emph{i.e.}, including DANN~\cite{ganin2016domain}, RTN~\cite{long2016unsupervised}, OSBP~\cite{saito2018open}, MDAN~\cite{zhao2018adversarial}, MDDA~\cite{zhao2020multi}, UAN~\cite{you2019universal}, DCTN~\cite{xu2018deep}, and UMAN~\cite{yin2022universal}. 
To ensure a fair comparison, we still the same evaluation metrics as those in the previous study~\cite{yin2022universal}, which represents the mean per-class accuracy over both the shared classes and the unknown class. 

Since the UFDA setting is fairly new in this field, we also compare the other setting HCLD$^2\star$ based on the proposed HCLD$^2$. Different from HCLD$^2$, HCLD$^2\star$ implements HCLD$^2$ with the Pseudo-Soft-Label (PSL) which is generated by averaging the output of source models, we weigh each class by the number of source models containing this class. 

\subsubsection{Implementation details.}   
In UFDA, the architecture in each node can be either identical or radically different. However, to ensure a fair comparison with previous UniMDA works, we maintain a common model architecture. Specifically, we utilize ResNet-50 as the backbone for all tasks. The projection head of the contrastive network is a 2-layer MLPs that outputs 128-dimensional embeddings. For model optimization, we employ stochastic gradient descent (SGD) training with a momentum of 0.9. The learning rate is decayed using the cosine schedule, starting from a high value (\emph{e.g.}, 0.005 for Office-31, Office-Home, and VisDA2017+ImageCLEF-DA) and decaying to zero. To follow the standard UniMDA training protocol, we use the same source and target samples, network architecture, learning rate, and batch size as in the UMAN~\cite{yin2022universal}. In decentralized training, the number of communication rounds $r$ plays a crucial role. To ensure a fair comparison with traditional UniMDA works, we adopt $r=1$ for all tasks. Furthermore, we implement all methods using PyTorch and conduct all experiments on an NVIDIA GeForce GTX 4*2080Ti, utilizing the default parameters for each method. 

\subsection{Experimental Results}
Here we present the comparison between our method and the above baseline methods. Some results are directly chosen from~\cite{feng2021kd3a}. From the results in Table~\ref{officehome_ofice31}, despite the raw data and model specifics are not available, HCLD$^2$ still can perform comparably across almost all tasks compared with the traditional UniMDA setting. It also shows that, although HCLD$^2\star$'s performance on VisDA+ImageCLEF-DA is not ideal, it achieves SOTA results across several tasks on Office-Home and Office-31. The results highlight the efficacy of our proposed HCLD$^2$ again and demonstrate the instability of directly using the soft outputs for the pseudo-label generation.  

\subsection{Ablation Study}
\subsubsection{Overall Component Effectiveness.}
We study the effectiveness of three key components (PHL Generation, GCLD, and MVD) in HCLD$^2$, with results shown in Table~\ref{ablation}. Results show that both GCLD and MVD significantly improved accuracy compared to the approach that removes MVD and GCLD only trains a classifier with the pseudo-labels (PHL or PSL).
By combining these two components we can obtain the best performance. 
Suffer from the one-hot setting, the method exclusively trains a classifier employing the PHL, resulting in consistently lower accuracy compared to the PSL. 
However, intriguingly, the integration of GCLD yields a remarkable outcome where the PHL-based approach significantly outperforms the PSL-based approach by a substantial margin. 

\subsubsection{Effectiveness of the PHL Generation.}
To further analyze the impact of different pseudo-label generated methods, we report the performance of HCLD$^2\star$ and HCLD$^2$ with varying settings of category in Table~\ref{class_setting}. We can see that HCLD$^2\star$ works better than HCLD$^2$ when the intersection of the multi-source label sets is non-empty. However, when the intersection is empty, the performance of HCLD$^2\star$ will suddenly decline along with the accuracy of PSL. On the other hand, HCLD$^2$ performs well with all category settings and shows a more stable performance compared with HCLD$^2\star$, which is sensitive to different category settings.   

\begin{table}[!t]
\centering
\setlength{\tabcolsep}{4.5pt}
\begin{tabular}{l l l l| c c c c }
    \hline
    \multicolumn{4}{c|}{\textbf{Methods}} & \multicolumn{4}{c}{\textbf{VisDA+ImageCLEF-DA}}\\
    \hline
    PSL &PHL &GCLD &MVD & C & I& P & Avg.\\
    \hline
    \hline
  \Checkmark & &   & &   56.5 & 52.1 & 53.2 & 53.9  \\
  \Checkmark & & & \Checkmark  & 60.3 & 59.5 & 55.0 & 58.3  \\
    \Checkmark & &\Checkmark & & 69.8  & 62.4 & 54.8 & 62.3  \\
  \Checkmark & & \Checkmark& \Checkmark & 77.3  & 67.5& 60.1 & 68.3   \\
   \hline
   & \Checkmark& & & 44.3  & 43.0 & 37.6 & 41.6  \\
   & \Checkmark&  & \Checkmark & 53.5 & 47.3 & 44.3 & 48.4 \\
    & \Checkmark&  \Checkmark&  & 74.5  & 70.0 & 55.2 & 66.6\\
   & \Checkmark& \Checkmark& \Checkmark &84.3 &73.5& 68.5 & 75.4  \\ 
  \hline
\end{tabular} 
\caption{Ablation Study. PSL: Pseudo-Soft-Label. PHL: Pseudo-Hot-Label. GCLD: Gaussian Mixture Model-based Contrastive Label Disambiguation. MVD: Mutual-Voting Decision.}
\label{ablation}
\end{table}
\begin{table}[!t]
\centering
\setlength{\tabcolsep}{5pt}
\begin{tabular}{l|c|c|c|c|c|c}
    \hline
    \textbf{UMDA-Matrix} & \textbf{M} & \multirow{1}*{\textbf{Ar}} & \multirow{1}*{\textbf{Cl}}& \multirow{1} *{\textbf{Pr}} & \multirow{1} *{\textbf{Re}} & \multirow{1}*{\textbf{Avg.}}\\
    \hline
    \hline
    \multirow{1}*{$\left[
 \begin{matrix}
   3&3&2&8 \\
   2&2&1&52 \\
  \end{matrix}\right]$} & \textbf{P}  & 36.4 & 47.3  & 55.7 & 66.8 & 51.5  \\
  &\textbf{S} & 48.8 & 50.6 & 65.9 & 76.3& 60.4  \\
  &\textbf{H}  & 68.9 & 51.9  & 72.6  & 80.5  & 68.5 \\
  \hline
  \multirow{1}*{$\left[
 \begin{matrix}
   4&3&3&10\\
   2&2&2&49\\
  \end{matrix}\right]$} & \textbf{P} & 47.9 & 43.5 & 57.7 & 65.5 & 53.6 \\
  &\textbf{S} & 59.9 &51.0 & 68.2& 76.2& 63.9 \\
  &\textbf{H} & 67.6 & 52.6& 69.9& 78.2& 67.1 \\
 \hline 
 \multicolumn{7}{c}{The intersection of MS: Empty $\uparrow$, Non-empty $\downarrow$} \\
 \hline
  \multirow{1}*{$\left[
  \begin{matrix}
   4&4&4&10\\
   2&2&2&50 \\
  \end{matrix}\right]$} &\textbf{P} & 68.3 & 52.7 & 72.4 & 75.4 & 67.2\\
  &\textbf{S} &80.3 & 62.3 &82.3 &88.9 & 78.4\\
  &\textbf{H} &77.4 &61.1 & 81.7 & 88.5 & 77.2\\
  \hline
 \multirow{1}*{$\left[
 \begin{matrix}
   6&6&6&10\\
   2&2&2&50\\
  \end{matrix}\right]$} &\textbf{P} & 75.0& 55.9 & 73.9& 82.8& 71.9 \\
  &\textbf{S} & 83.7 & 64.9&83.9 & 90.3& 80.7 \\
  &\textbf{H} & 81.4 & 62.1 & 80.9& 89.2& 78.4  \\ 
  \hline
\end{tabular} 
\caption{Comparison with different category settings on Office-Home measured by Accuracy (\%). MS: Multi-Source label sets. \textbf{M}: Methods. \textbf{P}: Pseudo-Soft-Label. \textbf{S}: Implementation of HCLD$^2$ with \textbf{P}. \textbf{H}: Our proposed HCLD$^2$.} 
\label{class_setting}
\end{table}

\subsubsection{Effectiveness of GCLD.} 
\begin{figure}[!t]
\centering
\subfloat[Accuracy]{
    \includegraphics[width=1.52in]{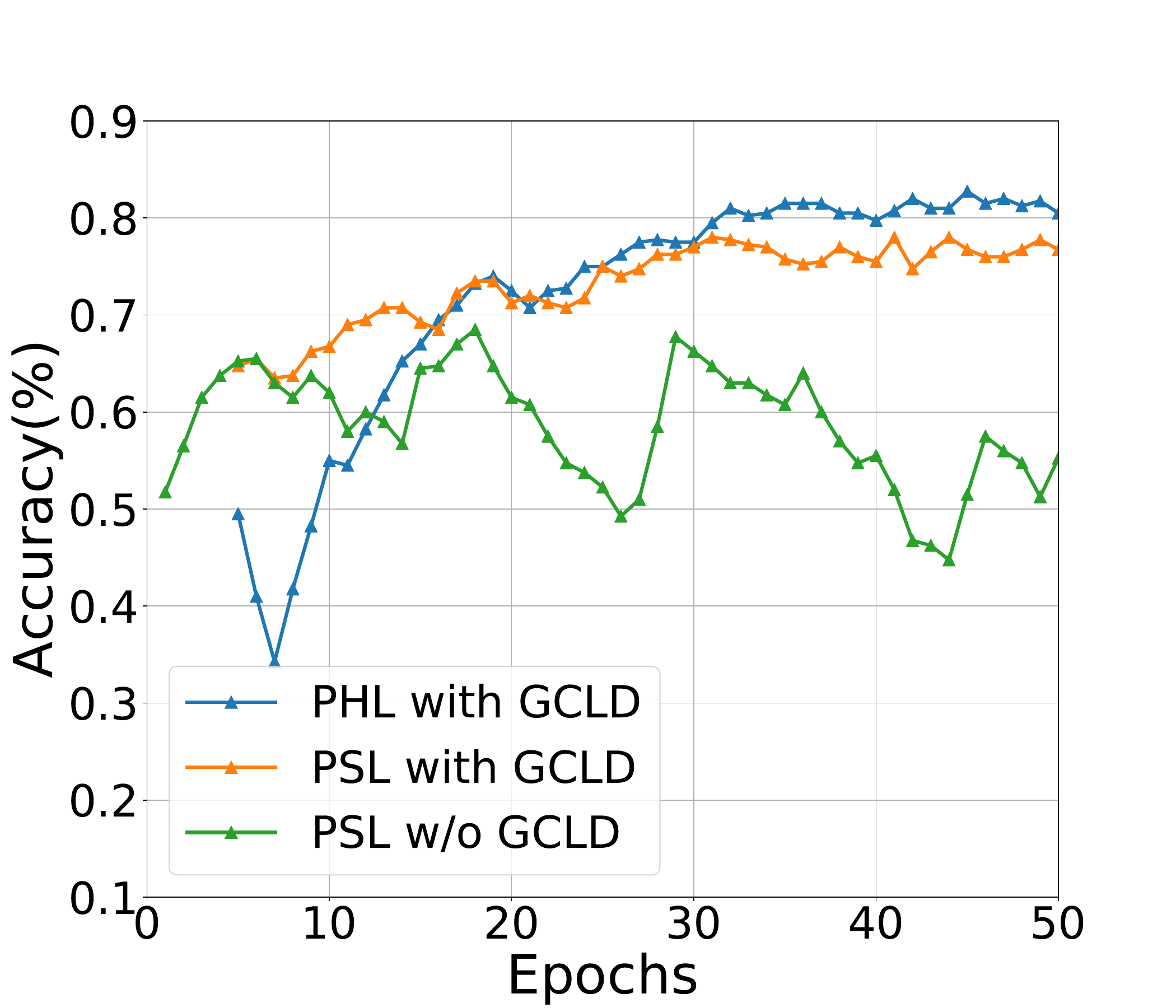}
    \label{C}}
    \hfil
    \subfloat[Sensitivity to $\lambda$]{
    \includegraphics[width=1.52in]{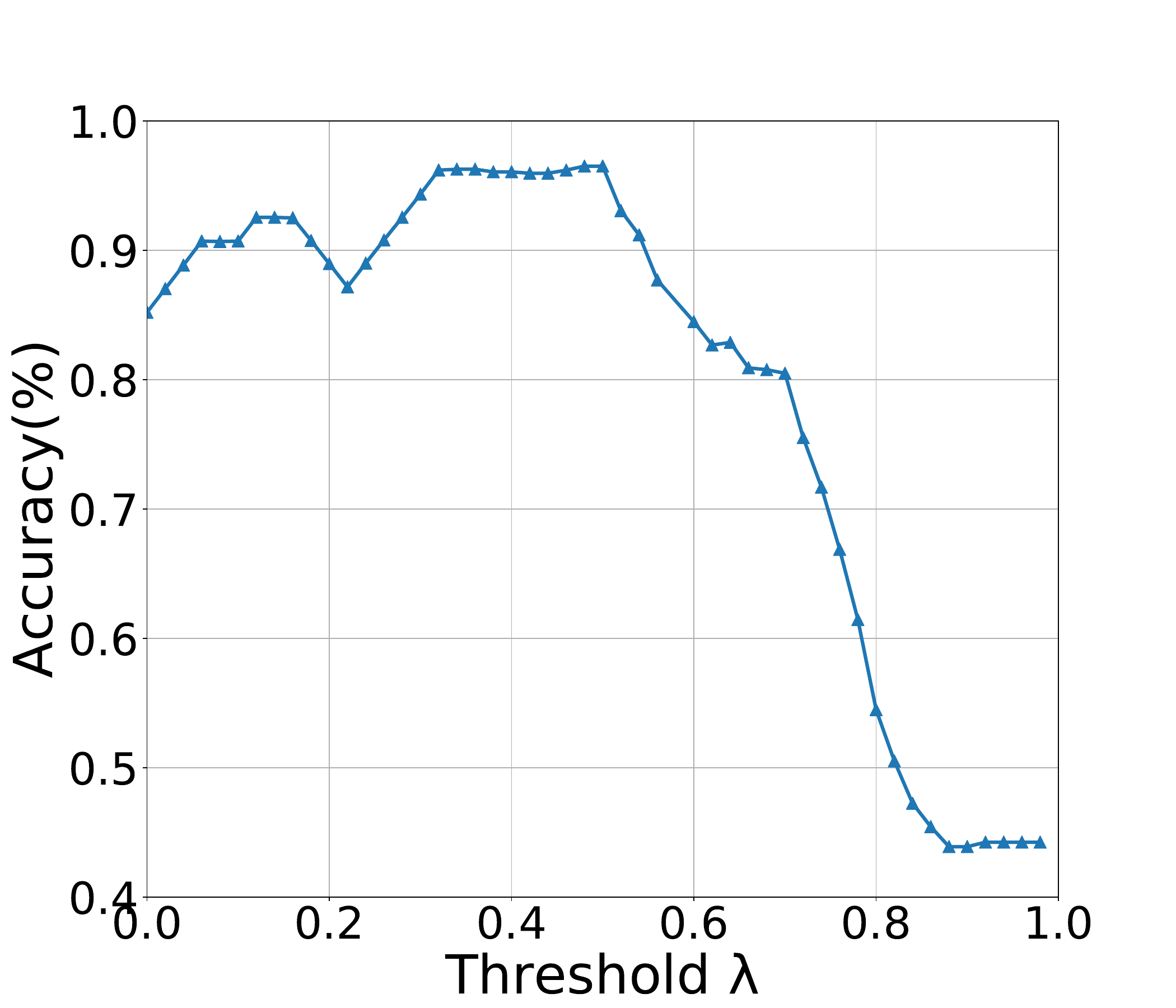}
    \label{voting}}
    \caption{(a) indicates the Accuracy of Pseudo-labels on task C in VisDA+ImageCLEF-DA. (b) plots the sensitivity to parameter $\lambda$ on task Dslr in Office-31. PSL: Pseudo-Soft-Label. PHL: Pseudo-Hot-Label. GCLD: Gaussian Mixture Model-based Contrastive Label Disambiguation. }
    \label{Label_disambiguation}
\end{figure}

In Figure.~\ref{C}, we report the performance of PSL with and without GCLD, and PHL with GCLD. As illustrated, PSL with GCLD outperforms the approach without GCLD by a large margin. In the initial epochs, PHL with GCLD may suffer from the one-hot setting. As the number of training epochs increases, PHL with GCLD will surpass the performance of PSL with GCLD.  

\subsubsection{Effectiveness of MVD.} 
In Table~\ref{voting}, we show results for the single-view and mutual-voting decision strategies. As a baseline, we implement HCLD$^2$ without incorporating any shared-class decision strategy. We establish the shared classes through voting outcomes within the source or target node for the single view. As we can see, MVD yields the most favorable results compared to any other single-view approach, although every single strategy exhibits improved performance over the baseline. Moreover, we study the parameter $\lambda$ on task Dslr. As shown in Figure.~\ref{voting}b, within a wide range of $\lambda$ (0.3-0.5), the performance only varies to a small degree, showing that our method is robust to different choices of $\lambda$.  
\begin{table}[!t]
\setlength{\tabcolsep}{5pt}
\centering
\begin{tabular}{l|c|c|c|c}
    \hline
    \textbf{Methodologies} &\textbf{A} & \textbf{D} & \textbf{W} & \textbf{Avg.}\\
    \hline
    \hline
    w/o MVD & 69.48 & 90.26 & 89.77 & 83.17 \\
    MVD (w/o target view) & 71.53& 94.52 & 92.43& 86.16   \\
    MVD (w/o source view)  & 75.18 & 96.55& 93.59 & 88.44 \\
    MVD  &75.66 & 97.22 & 94.95 & 89.28\\
    \hline
\end{tabular} 
\caption{Comparison with different shared-class decision strategy on Office-31 measured by Accuracy (\%). }
\label{voting}
\end{table}

\subsubsection{Why not Source-Free DA.} 
Although FDA and SFDA are similar to some extent (\emph{e.g.}, only the pre-trained source model is accessible to the target domain), they are essentially different. FDA has an important assumption, \emph{i.e.}, the decentralized source clients keep communicating the updated source black-box models during the training process, whereas this does not hold in SFDA at all. Both our proposed scenario UFDA and our method HCLD$^2$ heavily rely on this assumption and aim to make the black-model communication in a more practical condition. 
Indeed, the difference between the SFDA and FDA settings of our method could be reflected in Table~\ref{communication_r}. As seen, without black-box model communication in SFDA, the performance of our model significantly drops. Moreover, the FDA performance increases with the communication round $r$. 

\begin{table}[!htbp] 
\centering
\setlength{\tabcolsep}{5.5pt}
\begin{tabular}{c|c|ccccc }
    \hline
    & & \multicolumn{5}{c}{\textbf{FDA ($r$)}} \\
    \cline{3-7}\multirow{1}*{}& \textbf{SFDA} & 0.2  & 0.5 &1 & 5 & 10 \\
    \hline
    \hline
    A & 69.59& 72.76 & 75.09 &  75.66 & 77.39 & 77.72 \\
    D & 86.92& 92.3 &95.04 & 97.22 & 97.27 &97.11 \\
    W & 90.59& 92.49 & 94.34 & 94.95 & 95.54 & 96.93 \\
    \hline
    Avg. &83.03 &85.85 & 88.15 & 89.28 & 90.07 & 90.59 \\
  \hline
\end{tabular} 
\caption{Accuracies (\%) on Office-31 for SFDA and FDA with various communication rounds $r$ (per epoch).} 
\label{communication_r}
\end{table}

\section{Conclusion}
This work investigated a more practical scenario, UFDA, where we relax the comprehensive assumptions such as configuration specifics nor the prior label set overlap across multi-source and target domains as in most FDA scenarios.
We propose a new optimization methodology HCLD$^2$ to address UFDA and cluster-level strategy called MVD to distinguish shared and unknown classes during inference. 
Through extensive evaluations of three benchmark datasets, we demonstrate that HCLD$^2$ is capable of achieving comparable performance as conventional MDA baselines even with much less source knowledge. In the future, we may explore methods to further minimize additional assumptions (\emph{e.g.}, source label sets) in our UDFA, aiming for a more relaxed FDA scenario.
                    
\subsubsection{Acknowledgment}
This work was supported in part by the National Key R\&D Program of China 2022YFF0901800, in part by the NSFC Grant. (No.61832008, 62176205, and 62072367), in part by the Hong Kong Research Grants Council General Research Fund (17203023), in part by The Hong Kong Jockey Club Charities Trust under Grant 2022-0174, in part by the Startup Funding and the Seed Funding for Basic Research for New Staff from The University of Hong Kong, and part by the funding from UBTECH Robotics.

\bigskip
 
\bibliography{aaai24}
\end{document}